%% file: main.tex
\documentclass{llncs}

\usepackage{amssymb}
\usepackage{graphicx}

\usepackage{url}

\usepackage{multirow}
\usepackage{multicol}
\usepackage{amsmath}
\usepackage{amssymb}

\usepackage{graphicx}
\usepackage{color}
\usepackage[utf8]{inputenc}
\usepackage{hyperref}
\usepackage{verbatimbox}
\usepackage{multirow}
\usepackage[lofdepth,lotdepth]{subfig}

\usepackage{float}
\usepackage{amssymb}
\usepackage{graphicx}
\usepackage{color, colortbl}
\definecolor{Gray}{gray}{0.9}
\usepackage{url}
\newcommand\bx{\mathbf{x}}
\newcommand\btheta{{\boldsymbol{\theta}}}
\newcommand\bphi{{\boldsymbol{\phi}}}
\newcommand\bb{{\boldsymbol{b}}}

\usepackage{etoolbox}

\begin{document}


\author{Thomas Gerald\inst{1} \and Aurélia Léon\inst{1} \and Nicolas Baskiotis\inst{1}  \and Ludovic Denoyer\inst{1}}

\institute{Sorbonne Universités, UPMC Univ Paris 06, UMR 7606, LIP6,  Paris, France\\ \email{surname.name@lip6.fr} }

%
%



\title{Binary Stochastic Representations for Large Multi-class Classification}

\maketitle
\vspace{-15pt}

\begin{abstract}
Classification with a large number of classes is a key problem in machine learning and corresponds to many real-world applications like tagging of images or textual documents in social networks. If one-vs-all methods usually reach top performance in this context, these approaches suffer of a high inference complexity, linear w.r.t the number of categories. Different models based on the notion of binary codes have been proposed to overcome this limitation, achieving in a sublinear inference complexity. But they \textit{a priori} need to decide which binary code to associate to which category before learning using more or less complex heuristics. We propose a new end-to-end model which aims at simultaneously learning to associate binary codes with categories, but also learning to map inputs to binary codes. This approach called \textit{Deep Stochastic Neural Codes (DSNC)} keeps the sublinear inference complexity but do not need any \textit{a priori} tuning. Experimental results on different datasets show the effectiveness of the approach w.r.t baseline methods. 
\end{abstract}
\vspace{-25pt}

\input{intro.tex}

\vspace{-10pt}
\input{related.tex}

\vspace{-10pt}
\input{model.tex}
\vspace{-10pt}
\input{expe.tex}

\vspace{-7pt}
\bibliography{article}
\bibliographystyle{splncs}

\end{document}

%% file: intro.tex
\section{Introduction}
Classification problems involving very large number of classes have progressively emerged over the last years and are attracting an increased attention in the machine learning community 
(for instances challenges LSHTC \cite{Partalasa} or ImageNet \cite{ILSVRC15} with up to thousands of classes).
When facing such a large number of categories, one challenge is to keep the inference complexity as a reasonnable level: classical approaches have an inference complexity which is linear w.r.t the number of categories.
Concerning neural networks, this complexity is due to the last layer 
that computes one score for each category. If the use of GPUs can drastically reduce the computation time, the complexity still remains very high. Note that one versus all techniques are, up to now, among the strongest contender in terms of classification performances for large number of classes \cite{Perronnin2012b}.  

 In this paper, we propose a new deep neural model called \textit{Deep Stochastic Neural Codes (DSNC)} with a sublinear inference time thanks to a discrete binary hidden layer: an input is first mapped to a  small binary code and then a decoding process assigns
the corresponding label. The proposed model aims to learn simultaneously which code to associate with which category and how to map inputs to codes in an end-to-end manner.
The presented work is closely related to the field of binary hashing which use binary coding to index items (images, documents, \ldots). However, the goal differs largely : semantic hashing looks to preserve similarities 
between the projected representations; the objective of our model is to discover codes able to represent the latent organization of the classes.
 Therefore, contrarily to most existing neural approaches using continuous derivation and thresholding to learn the mapping, the proposed model integrated stochastic units to sample efficiently the code space. 
Since our architecture involves a discrete non-differentiable layer, we propose a learning algorithm based on the \textit{Straight Through} estimator proposed in \cite{BengioLC13}.
The contributions of the paper are thus threefolds: 1) we propose a new family of discrete deep neural network aiming at classifying when the number of categories is large by learning to map inputs to binary codes, and codes to categories;
2) we present an end-to-end learning algorithm that do not need any \textit{a priori} heavy work, the model being able to decide by itself which code to associate to which categories;
3) we show that this model is able to outperform existing techniques in term of accuracy while keeping a low inference complexity.
The paper is organized as follows: section \ref{relatedwork} presents the state of the art in multi-class classification and related work in binary hashing representation; section  \ref{model} presents the proposed model and the learning procedure;  section \ref{expes} presents the evaluation of the proposed model on usual large scale datasets and the analysis of the results.

%% file: related.tex
\section{Related Work}
\label{relatedwork}

One of the main issue in large scale multi-class classification is the trade-off between the prediction accuracy  and the time complexity for the classification of an example - the inference time with respect to $K$ the number of classes. 
The classical meta-algorithm \textit{one-versus-one} trains $O(K^2)$ classifiers to pairwise discriminate labels; \textit{one-versus-rest} trains $O(K)$ classifiers to distinguish each class from all the others. Both algorithms show efficient to deal with thousand classes  but at the price of at best an inference time which is linear with the number of classes. Both methods are thus prohibitive when considering a very large number of classes.
Different approaches have been proposed for reducing the complexity to a sublinear complexity w.r.t the number of categories. For example, especially when an existing hierarchy is available, one can use hierarchical models 
 \cite{bengio2010,Weston2013,puget2015hierarchical} classifying in logarithmic time.
When the structure of the output space is unknown (no class ontology), the state of the art approach is the Error Correcting Output Code approach (ECOC, \cite{Dietterich1995}): a binary code is associated to each category, and a function is learned
to map  any input to one possible code. 
Since defining binary codes of size $\log K$ is sufficient\footnote{In practice, a code of size $k \log K$ is needed with $k$ ranging between $10$ and $20$.} to encode $K$ categories, the resulting inference complexity will be $O(\log K)$. But those approaches suffer from two main drawbacks: (i) choosing which code to associate to which category is usually made by hand, even by using random codes or by using complex heuristics \cite{Zhong13,cisse2012} that need a heavy learning process. (ii) Even if codes are \textit{carefully} chosen, the performances is usually lower than classical one-vs-all approaches. Learning the mapping
 corresponds to multiple binary classification subtasks involving large number of classes. The ECOC performances are thus highly dependent on the separability of subsets of classes, which is know to be increasingly hardest with a growing number of classes.

On the other hand, mapping continuous representation to a binary one (known as hashing)  has been a topic of growing interest in  indexing large scale dataset. As large scale dataset contains an huge number of features  , performing a neighbors search to retrieve similar data requires an expensive computational time. Finding hashing functions from the initial description space to a lower binary one allow to perform a nearest neighbors query in sublinear time \cite{Norouzi14}.
Hashing algorithms can be divided in two categories: 
 Local Sensitive Hashing which use random projections of the data \cite{Gionis99,Weiss09}, and learning to hash algorithms which are data-driven, optimizing a loss function to preserve similarities \cite{Salakhutdinov09,Lai15,Do2016}.
 Those algorithms have shown impressive results for performance measures related to information retrieval as mean average precision \cite{Wang17}.
 However, they fail in classification tasks due to a poor recall rate: the hash functions are designed to preserve kind of metrics in the 
 hamming space but not to encourage discriminant codes between classes. 
 As noted by \cite{Liong15}, the compactness of the code is crucial: a larger code ensures a better precision measure and  less false positive, but at the same time decreases the recall and more false negative are retrieved.
These approaches are better to fragment original space than to perform  generalization  especially when the code length increases  \cite{Liong15}.

%% file: model.tex
\section{Deep Stochastic Neural Codes}
\label{model}

\subsubsection*{Model description}Let consider $\mathbb{R}^n$ and $\mathcal{K}=\{1,2,...,K\}$ the input  and the categories space respectively and $\mathcal{D} \in \mathbb{R}^n \times \mathcal{K}$ the training dataset. Let consider a size of code $c$ and the code space (or Hamming space) $\mathcal{B}=\{\bb \in \{0,1\}^c \}$,
and $b_i$ the i-th bit of a code $\bb$.

\begin{figure}[htbp]
	\centering
	\includegraphics[scale=0.4]{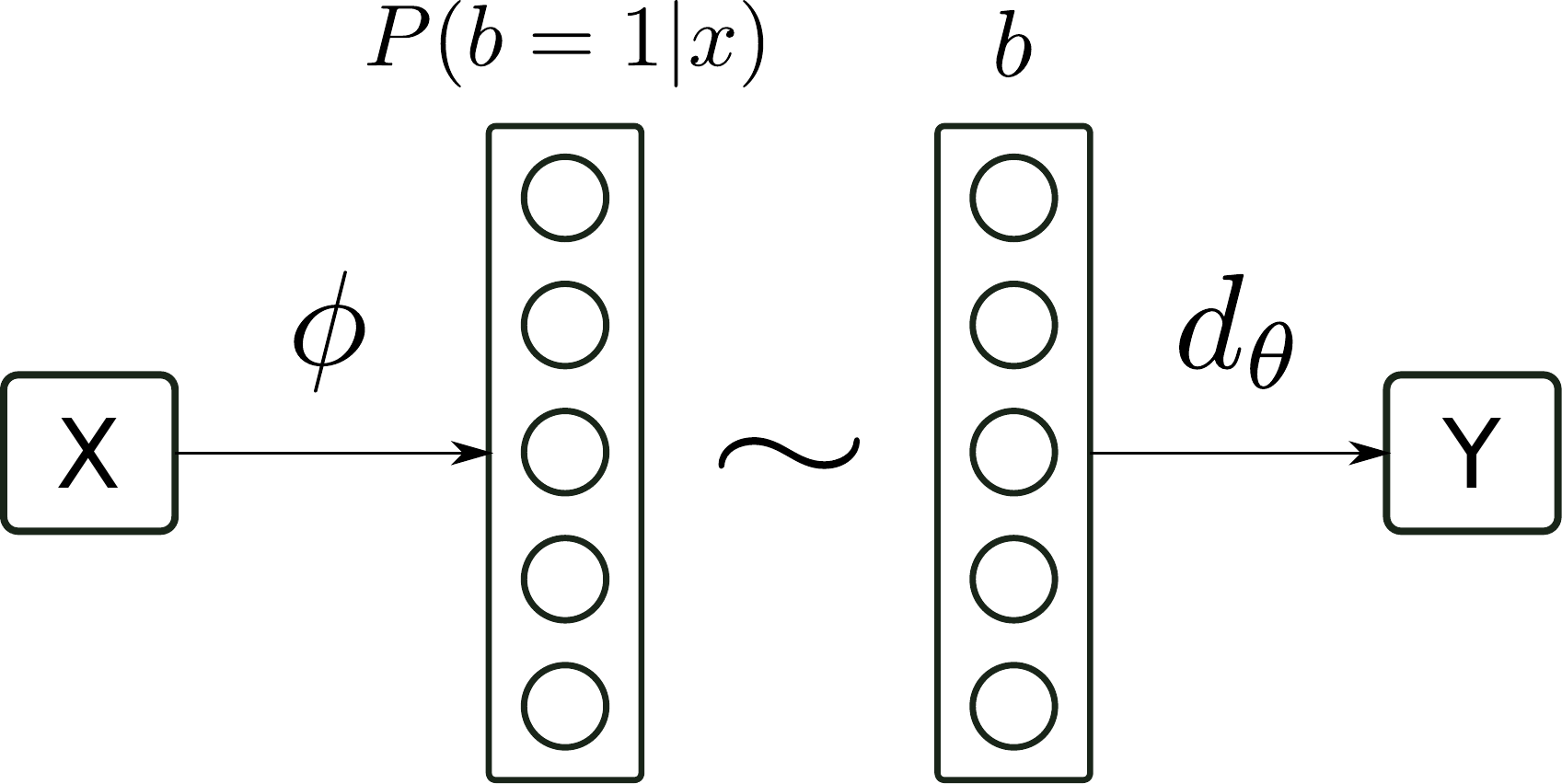}
    \caption{The DSNC model, with $\phi$ the probability distribution processed from input, $b$ the binary code drawn from the distribution and $d_\theta$ the decoding function from codes to classes.}
    \label{figmodel}
\end{figure}

Given an input $\bx \in \mathbb{R}^n$, the proposed model uses three different steps for a stochastic inference of the label as illustrated by Fig. \ref{figmodel}:  
\begin{itemize}
\item the first step maps the input $\bx$ to a probability distribution $P(\bx| x)$ over the binary codes noted $\phi(\bx)$;
\item the distribution is used in a second step to sample a code; 
\item the last step decodes the drawn code to a label.
\end{itemize}

To model the distribution $\phi$, we assume that the bits of a code are independents: each code bit $b_i$ can be modeled by a Bernoulli distribution of parameter noted $\phi_i(\bx)=P(b_i=1|\bx)$ and
the probability of a code given $\bx$ can be decomposed as $P(\bb|\bx)  = \prod_{i=1}^c \phi_i(\bx)^{b_i} (1-\phi_i(\bx))^{1-b_i}$.

Two functions needto be learned simultaneously:  $\phi : \mathbb{R}^n \to [0,1]^c$ which encodes the input to a code distribution; and the decoding function, noted $d_\theta : \{0,1\}^c\to \mathcal{K}$, which maps a category  to each code. 
 In the inference process instead of sampling we will choose directly the most probable value for each component in respect to $\phi(\bx)$.

Given a code, we propose two different decoding methods to infer the corresponding class. 
The first one consists in using a function trained during the learning phase to compute the probability of each category for a given code: we will refer this decoding function as \textit{linear-decoding}. 
The second one retrieves  the nearest neighbor of the  queried code among the codes encountered during the learning phase and outputs the associated class. We refer this decoding methods as \textit{nearest-neighbor-decoding}.

\subsubsection*{Learning procedure}
\label{learning}
Due to the stochastic sampling of the codes, the loss for a given couple $(\bx,y) \in \mathbb{R}^n\times \mathcal{K}$ is expressed as an expectation over the distribution of codes determined by $\phi(\bx)$ : 
$E_{\bb\sim\phi(\bx)} [\mathcal{L}(d_\theta(\bb),y)]$, with $\mathcal{L}$ an usual loss function (in the following we will use the negative likelihood as loss function, as it is usual in multi-class problems).
The optimization problem associated to the proposed model considering $\bb$ drawn from the distribution $\phi(\bx)$ can be written :

\begin{equation}
  \begin{aligned}
 \arg\min_{\phi,\theta}  J(\phi, \theta) = & E_{(\bx,y)}\left[ E_{\bb\sim \phi(\bx)} \left[ \mathcal{L}(d_\theta(\bb), y)  \right] \right]\\
                   = &\int ( P_{\phi}(\bb|\bx)\mathcal{L}(d_\theta(\bb),y) )P(y|\bb)P(\bx) d\bx dy d\bb
  \end{aligned}
\end{equation}


Optimizing this function using gradient descent algorithm requires an estimation of the gradient of $J(\phi, \theta)$ :

  \begin{flalign}
    \nabla  J(\phi, \theta) &=\int \nabla( P_{\phi}(\bb|\bx)\mathcal{L}(d_\theta(\bb),y) ))P(y|\bx)P(\bx) d\bx dy d\bb \\
     &= \int P_{\phi}(\bb|\bx) \nabla(log( P_{\phi}(\bb|\bx))) \nabla(\mathcal{L} (d_\theta(\bb),y) )P(y|\bx)P(\bx) d\bx dy d\bb \nonumber \\
     &  + \int \nabla( P_{\phi}(\bb|x)) (\mathcal{L} (d_\theta(\bb),y) )P(y|\bx)P(\bx) d\bx dy d\bb 
  \end{flalign}

A first approach to optimize this error function consist in using the REINFORCE algorithm
\cite{Williams1992}, a Monte-Carlo approximation of the gradient using  $M$ sampling over $\phi$ for each example:
\begin{equation}
	\nabla_{\phi,\theta} J(\phi,\theta) \approx  \frac{1}{\left|\mathcal{D}\right|} \sum \limits_{(x,y) \in \mathcal{D}}\left[\frac{1}{M} \sum \limits_{1}^{M} \nabla_{\phi,\theta}\left(log(\phi\left(\bx)\right)\right)\mathcal{L}(d_\theta(\bb^{\bx}),y) + \nabla_{\phi,\theta}\mathcal{L}\left(d_\theta(\bb^{\bx}),y\right) \right]
\end{equation}

The first term of the equation is relative to the update of the $\phi$ function and the second  term to the update of the $d_\theta$. This approximation is unbiased, however it involves a long learning time and does not scale well in a large action space.
Recent alternative methods have been developed to approximate such non-differentiable gradient problem. We propose to use the Straight-Through estimator (\textit{STE}, \cite{BengioLC13}) which reported great performances. The STE estimates the gradient over a hard threshold function
by considering this non-differentiable function as the identity function for the back-propagating
procedure: it is an approximation gradient computation that allows to back-propagate through a
single layer of such stochastic units, as clearly the sign of the derivative is coherent with the wanted
weights correction. The update of the parameters is produced as follows:



\begin{align} 
\btheta_{t+1} & = & \btheta_{t} - \sum \limits_{(\bx, y) \in \mathcal{D}}\nabla_{\btheta_t}\mathcal{L}(d_{\btheta_t}(\bb^\bx), y) \\
\bphi_{t+1} & = & \bphi_{t} - \sum \limits_{(\bx, y) \in \mathcal{D}} \nabla_{\bb^\bx}\mathcal{L}(d_{\btheta_t}(\bb^\bx), y) \nabla_{\bphi_t}(\phi(x)) 
\end{align}

\subsubsection*{Structured Binary Latent Space}
\label{reg}
The main objective of the model is to guarantee a latent code space able to generalize: through the learning process, several codes can be associated to a given class. However, 
to avoid the fragmentation of the space as in binary hashing, codes of a same class have to be close in the latent space. 
Toward this objective, we introduce a regularization term to minimize  the \textit{intra-class} and maximize the \textit{inter-class} distances. Considering the  two following sets : 

$$\mathcal{D}_{intra} = [ ((\bx, y), (\bx', y')) \in \mathcal{D }^2 | y = y' ]$$
$$\mathcal{D}_{inter} = [ ((\bx, y), (\bx', y')) \in \mathcal{D }^2 | y \neq y' ]$$
Thus the new objective function is: 
\begin{align}
J(\bphi, \btheta) = & E_{(\bx,y)}\left[ E_{\bb\sim \phi(\bx)} \left[ \mathcal{L}(d_\theta(b), y)  \right] \right] \nonumber \\
&+ \beta \sum \limits_{((\bx, y), (\bx', y'))\in \mathcal{D}_{intra}}  \|\phi(x)-\phi(x')\|^2  \label{intra} \\ 
  & - \gamma \sum \limits_{((\bx, y), (\bx', y'))\in \mathcal{D}_{inter}}  \|\phi(x)-\phi(x')\|^2 \label{inter} 
\end{align}

Where the minimization of the intra-class distance is represented by the first term \ref{intra} and the maximization of the inter-class distance by the second term \ref{inter} with  $\beta$, $\gamma$ the coefficients associated to each of those regularizations.
\subsubsection*{Complexity}
The complexity of the inference is essentially due to the decoding function. For the first investigated variant, the \textit{linear-decoding}, the complexity is the same as usual multi-class neural networks: the inference
takes $O(cK)$ operations to compute the $K$ probabilities of each class, linearly dependent to the number of classes.
Concerning the nearest neighbor decoding variant, finding with brute force the nearest neighbor has a time complexity in $O(kc)$ with $k$ the number of training codes. 
However, nearest neighbors search in hamming space is a well known problematic and hence sub-linear methods have been developed to face the problem.

For instance,  the proposed method in \cite{Norouzi14} achieves a complexity  in $O(\frac{c~ \sqrt[]{k}}{\log_2{k}})$.
However, for both methods, all codes can be stored in memory when the size of code is small. In this case, after the model training step, all possible codes are enumerated and decoded by one of the two decoding methods to associate them the corresponding class. The time complexity is constant and negligible in this case, but the space complexity is high in $O(2^c)$ to store the codes which prevent to use large codes.



%% file: expe.tex
\section{Experiments}
\label{expes}
This section presents the evaluation of the proposed model on three usual large scale datasets with a large numbers of classes (see Table \ref{dataset} for the detailed characteristics):
\begin{itemize}
\item \textit{ALOI} \cite{Galar:2013:DCS:2514172.2514280} is a dataset of 1k classes of sift features extracted from image objects; 
 \item \textit{DMOZ} dataset \cite{Partalasa} is composed of short text description preprocessed in a bag of word representation; this dataset contains $12275$ classes  and a large vocabulary input size.
In the evaluation, we use the complete dataset but also subsampled datasets with 1k classes which will be referred as \textit{DMOZ-1K}. 
\item The last dataset is \textit{ImageNet}  with 1k image categories. Instead of using raw images, the features from the pre-trained model \textit{resnet-152} are used as inputs \cite{He2015}.
\end{itemize}

The protocol setting is identical for all datasets: $80\%$ of data are randomly drawn to form the training set, $10\%$ to be used as a validation set and the last $10\%$ as a test set where
the accuracy is evaluated.
The experiments are conducted using the \textit{STE} gradient estimator.
and  using an adaptive gradient descent optimizer namely Adam \cite{DBLP:journals/corr/KingmaB14} using mini-batch from size 100 to 1000 samples each. 
In all experiments, the encoder is a linear function followed by a sigmoid activation and the decoder used to train the network a linear function followed by a \textit{softmax} activation. 

We evaluate the proposed model DSNC with the two decoding variants - the learned linear decoder, noted \textit{linear} and the nearest neighbor decoder, noted \textit{NN} -
and for both variants, we tested the regularization proposed in section \ref{reg}, noted \textit{Reg}, and without. Moreover we adapt the regularization factor during learning, in increasing the factor when ($\times 2$) the validation accuracy increase and decreasing the factor ($\times \frac{1}{2}$) when the validation accuracy decrease.

For selecting the best hyper-parameters we selected the best validation accuracy for the learning-rate and the initial values of the regularization factor.

\begin{figure}[!htpb]
 \centering
	\includegraphics[width=\linewidth]{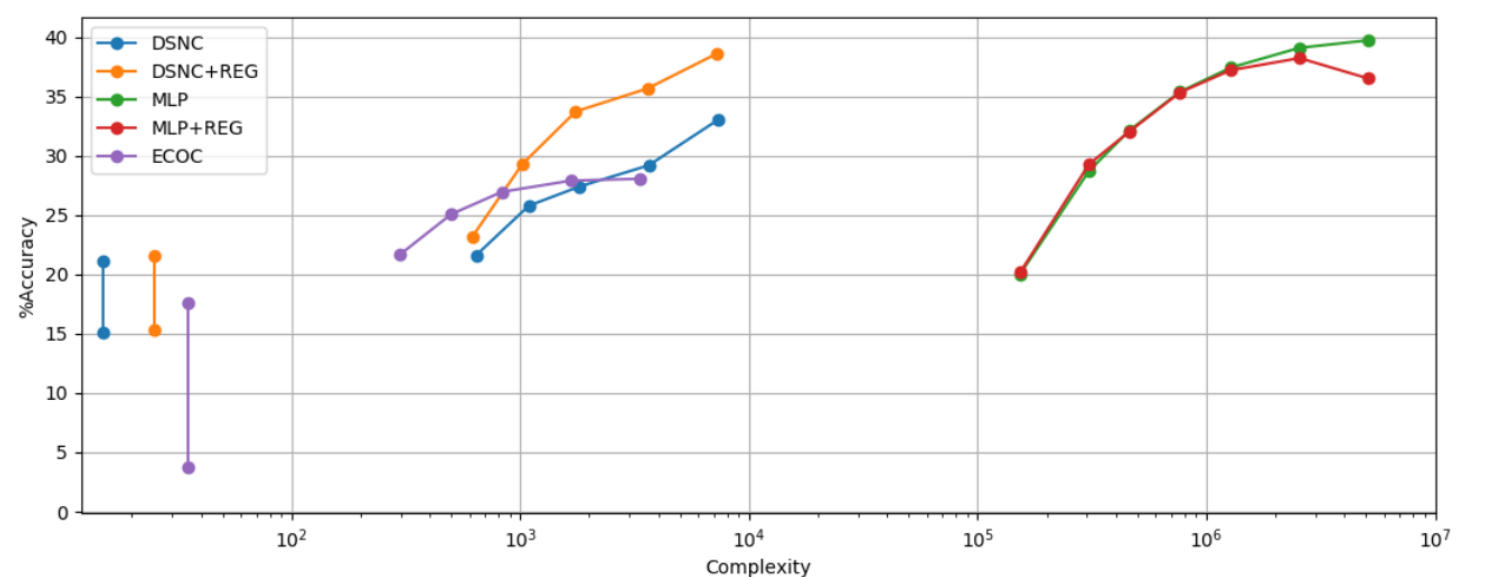}
    \caption{Complexity and Accuracy trade-off on DMOZ-12K}
    \label{DMOZComplexity}
\end{figure}

We compare the results to a classical multi-class multi-layer perceptron (\textit{MLP}) with a hidden layer fully-connected\footnote{The code size denotes in this case the number of hidden units.} and
to an ECOC algorithm with linear classifiers\footnote{One-versus-all algorithm has been tested with results  similar to the best MLP score.}.
The table \ref{accuracy} sums up the accuracy obtained in test and the Fig. \ref{DMOZComplexity} shows the accuracy w.r.t. the theoretical decoding t
ime achieved by DSNC with \textit{NN} decoder compared to ECOC and MLP models on the DMOZ-12K dataset\footnote{All the models share the same encoding complexity (fowarding the input to the hidden layer, discrete or not).} using the formulas reported  in section \ref{model}.
The ECOC and DSNC models on the left of the figure have a code size smaller than $30$ and thus have all the same complexity, a negligible constant decoding time as codes can be stored in memory. 
The model with the highest complexity - linear in the number of classes - are the MLP model (and our variant using the \textit{linear} decoder not represented in the figure).
The models with middle complexity, DSNC-NN and ECOC, are obtained by varying the code size between $30$ and $400$. The inference time is sub-linear w.r.t. the number of classes, the exact complexity depending on the trade-off between space and time complexity. On this dataset, the results show that our model outperforms ECOC models for a same complexity and is competitive with MLP for a great gain of complexity. 

\vspace{-25pt}
\begin{table}[htbp]
	\centering
    \caption{Characteristics of the datasets.}
    \begin{tabular}{|c|c|c|}
    \hline
    	Dataset Name & number of classes & number of examples \\
    \hline
    	DMOZ-1K& 1000 &$ 41,846 \pm 5,255$ \\
		DMOZ-12K& 12275 & $155,775$\\
        ALOI& 1000 & $108,000$ \\
        IMAGENET& 1000 & $14,197,122$\\
    \hline
	\end{tabular}
	\label{dataset}
\end{table}

\vspace{-10pt}

\begin{table}[htpb]
\begin{scriptsize}
     \caption{Accuracy of the proposed model DSNC and the two baselines on the different datasets. The gray background indicates a constant decoding time.
     }
	\centering
	\begin{tabular}{|c|c|c|c|c|c||c|c||c|}
\hline
	      &model & \multicolumn{4}{c||}{DSNC} & \multicolumn{2}{c||}{MLP} & ECOC\\
	      
    	\hline
		 dataset&code size& linear &NN & linear+Reg. & NN+Reg.&  & Reg.&  \\
        \hline
        \multirow{6}{*}{DMOZ-1K}&12&\cellcolor{Gray}31.772 &\cellcolor{Gray}22.425 &\cellcolor{Gray}33.156 & \cellcolor{Gray}23.492 & 39.204 & 39.716&\cellcolor{Gray}10.738 \\
        &24&\cellcolor{Gray}39.736 &  \cellcolor{Gray}36.779 & \cellcolor{Gray}41.326& \cellcolor{Gray}37.028 & 48.496 & 48.748&\cellcolor{Gray}22.144 \\ 

        &36& 42.928& 41.208& 45.414& 42.776& 51.48  & 51.716 & 27.589 \\ 
        &60& 46.84& 44.734& 48.742& 47.41& 53.532 & 54.084 & 33.850 \\ 
        &100& 49.164& 46.807& 50.984& 49.888 & 54.954 & 55.658 &38.212 \\ 
        &200& 51.058& 49.065& 53.052& 52.355 & 56.262& 56.908&41.33 \\ 
        \hline

        \multirow{6}{*}{DMOZ-12k}&12 &1\cellcolor{Gray}5.09 & \cellcolor{Gray}15.2&  \cellcolor{Gray}15.34&\cellcolor{Gray}15.24 & 20.05 &  20.18& \cellcolor{Gray}3.8\\ 
        &24 &  \cellcolor{Gray}21.15 & \cellcolor{Gray}18.79 & \cellcolor{Gray}21.64&\cellcolor{Gray}19.27 & 28.74&29.3& \cellcolor{Gray}17.591\\

        &36 & 24.83&22.21 & 25.95 &24.17&   32.14 & 32.05& 21.71 \\
        &60 & 28.36 &25.97& 29.71 &29.96& 35.41 & 35.36&  25.079\\
        &100 & 30.42&27.6& 31.98 &33.94 &  37.45 & 37.23& 27\\
        &200 & 32.22 &29.32& 33.95 &35.95 &  39.12 & 38.25 & 27.91\\
        &400 & 31.96 &33.02 & 33.6 &38.65 &36.71& 39.75 & 28.08 \\
        \hline




        \multirow{6}{*}{ALOI }&12 & \cellcolor{Gray}34.918  &\cellcolor{Gray}34.84 & \cellcolor{Gray}33.328&\cellcolor{Gray}33.366& 82.04&81.992& \cellcolor{Gray}1.53\\
        &24 & \cellcolor{Gray}67.92 &\cellcolor{Gray}63.66 & \cellcolor{Gray}66.064& \cellcolor{Gray}64.19 & 88.174 & 87.27&\cellcolor{Gray}4.21\\

        &36 & 76.73&74.19 & 75.84&79.94&  89.91& 88.18 & 5.81\\
        &60 & 83.478&83.19&82.66&81.74& 92.22 & 89.78 & 9.03\\
        &100 &88.014 &88.58&87.606&88.72 & 99.96 & 90.79 &13.8\\
        &200 & 91.288&91.88&90.542&91.26 & 95.15& 92.41&22.4 \\
        \hline


    	\multirow{6}{*}{IMAGENET} 
        &12 &\cellcolor{Gray}1.5& \cellcolor{Gray}0.82&  \cellcolor{Gray}1.49& \cellcolor{Gray}0.805 & 14.6 & 13.11& \cellcolor{Gray}12.32\\
        &24 &\cellcolor{Gray}15.6& \cellcolor{Gray}7.705 & \cellcolor{Gray}9.01&\cellcolor{Gray}4.595 & 53.19&53.46& \cellcolor{Gray}32.42 \\

        &36 & 53.82&36.46 & 45.74 &25.14&59.11& 58.85&  45.03\\
        &60 & 60.07&48.61& 63 &53.655& 60.83 & 63.66&  56.5\\
        &100 & 68.66&63.405& 68.81 &63.515& 65.9 &66.81 &64.5\\
        &200 & 70.67&66.935& 71.61 &68.54& - & 67.3&69.76 \\
        \hline
	\end{tabular}

 	\label{accuracy}
\end{scriptsize}
\end{table}




Looking into details the results of Table \ref{accuracy}, it is remarkable that the results of the NN decoder are very close to the linear decoder: it is an indication of the generalizing ability of the learned code space. Our model outperforms the ECOC baseline and in
 the best setting is very close to the robust MLP baseline. The proposed regularization improves the performances of our model  in most cases especially for large code size;
 in the case of the MLP, the regularization seems to have no effect. 
For the maximal inference speed-up (constant time prediction), using very short codes, the results are degraded but our model outperforms clearly the \textit{ECOC} approach.
When considering larger codes size, the regularized DSNC-NN outperforms in most settings all the other approach with a  lower complexity. The proposed model successes to get a good trade-off between  accuracy - higher than \textit{ECOC} methods - and time complexity - better than the tested models.

The table \ref{intra_dist_dmoz} summarizes the distance intra-class and inter-class for the proposed model, with and without regularization for the codes of the training and the test sets.
The results show that the regularization has a real impact on the learned latent space: 1) the regularization decreases the distance intra-class and  increases the distance inter-class: this is an important feature to improve the nearest neighbor decoding as it allows better separation of codes of different classes; 2) the number of codes decreases with the introduction of the regularization, which indicates that the latent space is less fragmented. Moreover, less codes allows to speed-up the nearest neighbor decoding.  
To conclude, the experiments show that the regularization  favors the learning of fewer and more compact codes improving the performances of the model.

\vspace*{-15pt}
\begin{table}[H]
\begin{scriptsize}
    \caption{Latent space characteristics on DMOZ-1K dataset}
	\centering
	\begin{tabular}{|c|c|cc|cc|cc|cc|}
    	\hline
		 \multirow{2}{*}{Corpus} & \multirow{2}{*}{Distance}& \multicolumn{2}{c}{24} & \multicolumn{2}{c}{60} & \multicolumn{2}{c}{100} &\multicolumn{2}{c|}{200} \\ 
        \cline{3-10}
	  & & Reg & No Reg  & Reg & No Reg& Reg & No Reg& Reg & No Reg\\
        \hline
\multirow{3}{*}{train}&        intra-class  & $1.1 \pm 1$ & $ 2.3 \pm 1 $  &   $2.8 \pm 3 $& $10.9 \pm 3$ &   $ 8.82 \pm 6 $& $23 \pm 5$ & $ 18.4 \pm 11$ & $ 59.3\pm 10$\\ 
&       	 inter-class & $11.78 \pm 0$& $ 11.6  \pm 0 $ & $29.4 \pm 0$ & $26.6 \pm 0$ &$47 \pm 1 $& $40.8 \pm 1$ & $94.4 \pm 2$& $72.3 \pm 2$\\
 &      	 $\#$ codes & $8k$ & $15k$ &$15k$ & $28k$ & $24k$ & $29k$& $26k$ & $30k$\\
        \hline
\multirow{2}{*}{test}&      intra-class  &$4.8 \pm 3$ &$ 5.9 \pm 3 $ & $12.1 \pm 6$& $16.6 \pm 6$ & $20.1 \pm 10$ &$29.6 \pm 9$ &$37 \pm 19$& $63.9 \pm 17$\\ 
&       	inter-class&$10.8 \pm 1.$ &$10.5  \pm 1. $  &$27.0 \pm 2$   &$25.6 \pm 2 $& $45.0 \pm 3$& $41.6 \pm 4$ & $90.8 \pm 5$& $80.6 \pm 9$\\
        \hline
	\end{tabular} 

    \label{intra_dist_dmoz}
\end{scriptsize}
\end{table}

\section{Conclusion and perspectives}
 The presented model is a stochastic neural network architecture for multi-class classification, which learns jointly a function to map stochastically an input to a binary code and a decoder function associating codes to  classes.
The stochastic mapping between the input space and the latent binary space allows to explore efficiently the code space but introduces a non-differentiable layer. 
A Straigh-Through estimator is used to approximate the gradient and to learn the parameters.  In addition, a regularization is proposed to achieve a better structure of the latent space, with fewer and more compact codes. Thanks to the finite discrete property of the latent space, the proposed model is able to retrieve the class associated to each code with a constant negligible time for small code size and in the generic case  with a sublinear time w.r.t. to the number of classes. Experiments show the benefits of our model in terms of accuracy and time complexity.
The presented work is thus a first step toward learning binary latent space in large multi-class classification context. Further investigations concerns mainly the adaptation of the model to multi-class multi-label context - in which an example can be tagged by multiple label -  and the exploitation/analysis of the learned latent space for other classification tasks as automatic discovery of new classes and zero-shot learning.

\subsubsection*{Acknowledgments.}
This publication is based upon work supported by the King Abdullah University of Science and Technology (KAUST) Office of Sponsored Research (OSR) under Award No. OSR-2015-CRG4-2639.